\documentclass[preprint,12pt]{elsarticle}

\usepackage{listings}
\usepackage{color}

\definecolor{dkgreen}{rgb}{0,0.6,0}
\definecolor{gray}{rgb}{0.5,0.5,0.5}
\definecolor{mauve}{rgb}{0.58,0,0.82}

\usepackage{tikz}
\usetikzlibrary{backgrounds}
\usepackage{graphicx}
\usepackage{bm}
\usepackage{amsmath}
\usepackage{setspace}
\usepackage{hyperref}
\usepackage{subfigure}
\usepackage{multirow}
\hypersetup{
    colorlinks=true,       % false: boxed links; true: colored links
}
\usepackage{amssymb}
\usepackage{lineno}
\usepackage{diagbox}

\biboptions{sort&compress}

\journal{Journal Name}

\begin{document}

\begin{frontmatter}


\title{Forward-Backward Stochastic Neural Networks: Deep Learning of High-dimensional Partial Differential Equations}



% \author{Maziar Raissi$^{1}$, Paris Perdikaris$^{2}$, and George Em Karniadakis$^{1}$}
% \address{$^{1}$Division of Applied Mathematics, Brown University,\\ Providence, RI, 02912, USA\\
% $^{2}$Department of Mechanical Engineering and Applied Mechanics,\\ University of Pennsylvania,\\ Philadelphia, PA, 19104, USA}

\author{Maziar Raissi}
\address{Division of Applied Mathematics, Brown University,\\ Providence, RI, 02912, USA}

\begin{abstract}
Classical numerical methods for solving partial differential equations suffer from the curse dimensionality mainly due to their reliance on meticulously generated spatio-temporal grids. Inspired by modern deep learning based techniques for solving forward and inverse problems associated with partial differential equations, we circumvent the tyranny of numerical discretization by devising an algorithm that is scalable to high-dimensions. In particular, we approximate the unknown solution by a deep neural network which essentially enables us to benefit from the merits of automatic differentiation. To train the aforementioned neural network we leverage the well-known connection between high-dimensional partial differential equations and forward-backward stochastic differential equations. In fact, independent realizations of a standard Brownian motion will act as training data. We test the effectiveness of our approach for a couple of benchmark problems spanning a number of scientific domains including Black-Scholes-Barenblatt and Hamilton-Jacobi-Bellman equations, both in 100-dimensions.
\end{abstract}

\begin{keyword}
forward-backward stochastic differential equations, Black-Scholes equations, Hamilton-Jacobi-Bellman equations, stochastic control, deep learning, automatic differentiation
\end{keyword}

\end{frontmatter}

% \linenumbers

\section{Introduction}
Since their introduction \cite{bismut1973conjugate,pardoux1990adapted}, backward stochastic differential equations have found many applications in areas like stochastic control, theoretical economics, and mathematical finance. They have received considerable attention in the literature and interesting connections to partial differential equations have been obtained (see e.g., \cite{cheridito2007second} and the references therein). The key feature of backward stochastic differential equations is the random terminal condition that the solution is required to satisfy. These equations are referred to as forward-backward stochastic differential equations, if the randomness in the terminal condition is coming from the state of a forward stochastic differential equation. The solution to a forward-backward stochastic differential equation can be written as a deterministic function of time and the state process. Under suitable regularity assumptions, this function can be shown to be the solution of a parabolic partial differential equation \cite{cheridito2007second}. A forward-backward stochastic differential equation is called uncoupled if the solution of the backward equation does not enter the dynamics of the forward equation and coupled if it does. The corresponding parabolic partial differential equation is semi-linear in case the forward-backward stochastic differential equation is uncoupled and quasi-linear if it is coupled.\\

In this work, we approximate the aforementioned deterministic function of time and space by a deep neural network. This choice is inspired by modern techniques for solving forward and inverse problems associated with partial differential equations, where the unknown solution is approximated either by a neural network \cite{raissi2017physics_I,raissi2017physics_II,raissi2018deep} or a Gaussian process \cite{raissi2018numerical,raissi2018hidden,raissi2017inferring,raissi2017machine}. Moreover, putting a prior on the solution is fully justified by the similar approach pursued in the past century by classical methods of solving partial differential equations such as finite elements, finite differences, or spectral methods, where one would expand the unknown solution in terms of an appropriate set of basis functions. However, the classical methods suffer from the curse of dimensionality mainly due to their reliance on spatio-temporal grids. In contrast, modern techniques avoid the tyranny of mesh generation, and consequently the curse of dimensionality, by approximating the unknown solution with a neural network or a Gaussian process. Moreover, unlike the state of the art deep learning based algorithms for solving high-dimensional partial differential equations \cite{beck2017machine,weinan2017deep,han2017overcoming}, our algorithm (upon a single round of training) results in a solution function that can be evaluated anywhere in the space-time domain, not just at the initial point.

\section{Problem Setup and Solution methodology}
We consider coupled forward-backward stochastic differential equations of the general form
\begin{equation}\label{eq:FBSDE}
\begin{array}{l}
dX_t = \mu(t, X_t, Y_t, Z_t)dt + \sigma(t, X_t, Y_t)dW_t, ~~~ t \in [0,T],\\
X_0 = \xi,\\
dY_t = \varphi(t, X_t, Y_t, Z_t)dt + Z_t'\sigma(t,X_t,Y_t)dW_t, ~~~ t \in [0,T),\\
Y_T = g(X_T),
\end{array}
\end{equation}
where $W_t$ is a vector-valued Brownian motion. A solution to these equations consists of the stochastic processes $X_t$, $Y_t$, and $Z_t$. It is shown in \cite{antonelli1993backward} and \cite{ma1994solving} (see also \cite{pardoux1999forward,delarue2006forward,cheridito2007second}) that coupled forward-backward stochastic differential equations \eqref{eq:FBSDE} are related to quasi-linear partial differential equations of the form
\begin{equation}\label{eq:PDE}
u_t = f(t,x,u,Du,D^2u),
\end{equation}
with terminal condition $u(T,x) = g(x)$, where $u(t,x)$ is the unknown solution and
\begin{equation}
f(t,x,y,z,\gamma) = \varphi(t,x,y,z) - \mu(t,x,y,z)'z - \frac{1}{2}\text{Tr}[\sigma(t,x,y)\sigma(t,x,y)'\gamma].
\end{equation}
Here, $Du$ and $D^2u$ denote the gradient vector and the Hessian matrix of $u$, respectively. In particular, it follows directly from Ito's formula (see e.g., \cite{cheridito2007second}) that solutions of equations \eqref{eq:FBSDE} and \eqref{eq:PDE} are related according to 
\begin{equation}\label{eq:key}
Y_t = u(t, X_t), ~ \text{and} ~ Z_t = D u(t, X_t).
\end{equation}
Inspired by recent developments in \emph{physics-informed deep learning} \cite{raissi2017physics_I,raissi2017physics_II} and \emph{deep hidden physics models} \cite{raissi2018deep}, we proceed by approximating the unknown solution $u(t,x)$ by a deep neural network. We obtain the required gradient vector $Du(t,x)$ by applying the chain rule for differentiating compositions of functions using automatic differentiation \cite{baydin2015automatic}. It is worth emphasizing that automatic differentiation is different from, and in several respects superior to, numerical or symbolic differentiation; two commonly encountered techniques of computing derivatives. In its most basic description \cite{baydin2015automatic}, automatic differentiation relies on the fact that all numerical computations are ultimately compositions of a finite set of elementary operations for which derivatives are known. Combining the derivatives of the constituent operations through the chain rule gives the derivative of the overall composition. This allows accurate evaluation of derivatives at machine precision with ideal asymptotic efficiency and only a small constant factor of overhead. In particular, to compute the derivatives involved in equation \eqref{eq:key} we rely on Tensorflow \cite{abadi2016tensorflow} which is a popular and relatively well documented open source software library for automatic differentiation and deep learning computations.\\

\noindent Parameters of the neural network representing $u(t,x)$ can be learned by minimizing the following loss function given explicitly in equation \eqref{eq:loss} obtained from discretizing the forward-backward stochastic differential equation \eqref{eq:FBSDE} using the standard Euler-Maruyama scheme. To be more specific, let us apply the Euler-Maruyama scheme to the set of equations \eqref{eq:FBSDE} and obtain
\begin{equation}\label{eq:Euler}
\begin{array}{l}
X^{n+1} \approx X^n + \mu(t^n,X^n,Y^n,Z^n)\Delta t^n + \sigma(t^n,X^n,Y^n)\Delta W^n,\\
Y^{n+1} \approx Y^n + \varphi(t^n,X^n,Y^n,Z^n)\Delta t^n + (Z^n)'\sigma(t^n,X^n,Y^n)\Delta W^n,
\end{array}
\end{equation}
for $n = 0, 1, \ldots, N-1$, where $\Delta t^n := t^{n+1} - t^n = T/N$ and $\Delta W^n \sim \mathcal{N}(0, \Delta t^n)$ is a random variable with mean $0$ and standard deviation $\sqrt{\Delta t^n}$. The loss function is then given by
\begin{eqnarray}\label{eq:loss}
\sum_{m=1}^M \sum_{n=0}^{N-1} |Y^{n+1}_m - Y^n_m - \Phi^n_m \Delta t^n - (Z^n_m)' \Sigma^n_m \Delta W_m^n|^2 + \sum_{m=1}^M |Y^N_m - g(X^N_m)|^2,
\end{eqnarray}
which corresponds to $M$ different realizations of the underlying Brownian motion. Here, $\Phi^n_m := \varphi(t^n,X^n_m,Y^n_m,Z^n_m)$ and $\Sigma^n_m := \sigma(t^n,X^n_m,Y^n_m)$. The subscript $m$ corresponds to the $m$-th realization of the underlying Brownian motion while the superscript $n$ corresponds to time $t^n$. It is worth recalling from equations \eqref{eq:key} that $Y_m^n = u(t^n, X_m^n)$ and $Z_m^n = D u(t^n, X_m^n)$, and consequently the loss \eqref{eq:loss} is a function of the parameters of the neural network $u(t,x)$. Furthermore, from equation \eqref{eq:Euler} we have
\[
X^{n+1}_m = X^n_m + \mu(t^n,X^n_m,Y^n_m,Z^n_m)\Delta t^n + \sigma(t^n_m,X^n_m,Y^n_m)\Delta W^n_m,
\]
and $X^0_m = \xi$ for every $m$.

\section[Related Work]{Related Work\protect\footnote{This section can be skipped in the first read.}}
In \cite{weinan2017deep,han2017overcoming}, the authors consider uncoupled forward-backward stochastic differential equations of the form
\begin{equation}\label{eq:FBSDE_Arnulf}
\begin{array}{l}
dX_t = \mu(t, X_t)dt + \sigma(t, X_t)dW_t, ~~~ t \in [0,T],\\
X_0 = \xi,\\
dY_t = \varphi(t, X_t, Y_t, Z_t)dt + Z_t'\sigma(t,X_t)dW_t, ~~~ t \in [0,T),\\
Y_T = g(X_T),
\end{array}
\end{equation}
which are subcases of the coupled equations \eqref{eq:FBSDE} studied in the current work. The above equations are related to the semilinear and parabolic class of partial differential equations
\begin{equation}\label{eq:PDE_Arnolf}
u_t = \varphi(t,x,Du,D^2u) - \mu(t,x)'Du - \frac{1}{2}\text{Tr}[\sigma(t,x)\sigma(t,x)'D^2u].
\end{equation}
The authors of \cite{weinan2017deep,han2017overcoming} then devise an algorithm to compute $Y_0 = u(0,X_0) = u(0,\xi)$ by treating $Y_0$ and $Z_0 = Du(0,\xi)$ as parameters in their model. Then, they employ the Euler-Maruyama scheme to discretize equations \eqref{eq:FBSDE_Arnulf}. Their next step is to approximate the functions $Du(t^n,x)$ for $n=1,\ldots,N-1$ at time steps $t^n$ by $N-1$ different neural networks. This enables them to approximate $Z^n = Du(t^n,X^n)$ by evaluating the corresponding neural network at time $t^n$ at the spatial point $X^n$. Moreover, no neural networks are employed in \cite{weinan2017deep,han2017overcoming} to approximate the functions $u(t^n,x)$. In fact $Y^n = u(t^n,X^n)$ is computed by time marching using the Euler-Maruyama scheme used to discretize equations \eqref{eq:FBSDE_Arnulf}. Their loss function is then given by
\begin{eqnarray}\label{eq:loss_Arnulf}
\sum_{m=1}^M |Y^N_m - g(X^N_m)|^2,
\end{eqnarray}
which tries to match the terminal condition. The total set of parameters are consequently given by $Y_0$, $Z_0$, and the parameters of the $N-1$ neural networks used to approximate the gradients. There are a couple of major drawbacks associated with the method advocated in \cite{weinan2017deep,han2017overcoming}.\\

The first and the most obvious drawback is that the number of parameters involved in their model grows with the number of points $N$ used to discretized time. This is prohibitive specially in cases where one would need to perform long time integration (i.e., when the final time $T$ is large) or in cases where it is a requirement to employ smaller time step size $\Delta t$ in order to increase the accuracy of the Euler-Maruyama scheme. The second major drawback is that the method as outlined in \cite{weinan2017deep, han2017overcoming} is designed in such a way that it is only capable of approximating $Y_0 = u(0,X_0) = u(0,\xi)$. This means that in order to obtain an approximation to $Y_t = u(t,X_t)$ at a later time $t >0$, they will have to retrain their algorithm. The third drawback is that assuming $Y_0$ and $Z_0$ to act as parameters of the models in addition to approximating the gradients by $N-1$ distinct (not sharing any parameters) neural networks seems a little bit ad-hoc.\\

In contrast, the method proposed in the current work circumvents all of the drawbacks mentioned above by placing a neural network directly on the object of interest, the unknown solution $u(t,x)$. This choice is justified by the similar well-established approach taken by the classical methods of solving partial differential equations, such as finite elements, finite differences, or spectral methods, where one would expand the unknown solution $u(t,x)$ in terms of an appropriate set of basis functions. In addition, modern methods for solving forward and inverse problems associated with partial differential equations approximate the unknown solution $u(t,x)$ by either a neural network \cite{raissi2017physics_I,raissi2017physics_II,raissi2018deep,raissi2018multistep} or a Gaussian process \cite{raissi2018numerical,raissi2018hidden,raissi2017inferring,raissi2017machine,raissi2017parametric,perdikaris2017nonlinear,raissi2016deep}. The classical methods suffer from the curse of dimensionality mainly due to their reliance on spatio-temporal grids. Here, inspired by the aforementioned modern techniques, we avoid the curse of dimensionality by approximating $u(t,x)$ with a neural network. It should be highlighted that the number of parameters of the neural network we use to approximate $u(t,x)$ is independent of the number of the number of points $N$ needed to discretized time (see equation \eqref{eq:Euler}). Moreover, upon a single round of training, the neural network representing $u(t,x)$ can be evaluated anywhere in the space-time domain, not just at the initial point $u(0,X_0)$. Furthermore, we compute the required gradients $Du(t,x)$ by differentiating the neural network representing $u(t,x)$ using automatic differentiation. Consequently, the networks $Du(t,x)$ and $u(t,x)$ share the same set of parameters. This is fully justified by the theoretical connection (see equation \eqref{eq:key}) between solutions of forward-backward stochastic differential equations and their associated partial differential equations. A major advantage of the approach pursued in the current work is the reduction in the number of parameters employed by our model, which helps the algorithm generalize better during test time and consequently mitigate the well-known over-fitting problem.\\

In \cite{beck2017machine}, a follow-up work on \cite{weinan2017deep,han2017overcoming}, the authors extend their framework to fully-nonlinear second-order partial differential equations of the general form
\begin{equation}
u_t = f(t,x,u,Du,D^2 u),
\end{equation}
with terminal condition $u(T,x) = g(x)$. Here, let $X_t$ denote a high-dimensional stochastic process satisfying the forward stochastic differential equation
\begin{equation}\label{eq:FSDE}
\begin{array}{l}
dX_t = \mu(X_t)dt + \sigma(X_t) dW_t,\\
X_0 = \xi,
\end{array}
\end{equation}
where $\mu(X_t)$ is the drift vector and $\sigma(X_t)$ is the diffusion matrix. It then follows directly from Ito's formula \cite{cheridito2007second} that the processes
\begin{equation}\label{eq:processes}
\begin{array}{l}
Y_t := u(t, X_t),\\
Z_t := D u(t, X_t),\\
\Gamma_t := D^2 u(t, X_t),\\
A_t := \mathcal{L}Du(t,X_t) := D u_t(t, X_t) + \frac{1}{2} D\text{Tr} [D^2 u(t, X_t)\sigma(X_t)\sigma(X_t)^T],
\end{array}
\end{equation}
solve the second-order backward stochastic differential equation
\begin{eqnarray}\label{eq:2BSDE}
\begin{array}{l}
d Y_t = f(t, X_t, Y_t, Z_t, \Gamma_t)dt + \frac{1}{2}\text{Tr}[\Gamma_t\sigma(X_t)\sigma(X_t)^T] d t + Z_t^T d X_t,\\
d Z_t = A_t d t + \Gamma_t d X_t,\\
Y_T = g(X_T).
\end{array}
\end{eqnarray}
Similar to their prior works \cite{weinan2017deep,han2017overcoming}, the authors then devise an algorithm to compute $Y_0 = u(0,X_0) = u(0,\xi)$ by treating $Y_0$, $Z_0 = Du(0,\xi)$, $\Gamma_0 = D^2u(0,\xi)$, and $A_0 = \mathcal{L}Du(0,\xi)$ as parameters of their model. Then, they proceed by discretizing equations \eqref{eq:2BSDE} by the Euler-Maruyama scheme. Their next step is to approximate the functions $D^2u(t^n,x)$ and $\mathcal{L}Du(t^n,x)$ for $n=1,\ldots,N-1$, corresponding to each time step $t^n$, by $2(N-1)$ distinct neural networks. This enables them to approximate $\Gamma^n = D^2u(t^n,X^n)$ and $A^n = D^2u(t^n,X^n)$ by evaluating the corresponding neural networks at $X^n$. Moreover, no neural networks are employed in \cite{beck2017machine} to approximate the functions $u(t^n,x)$ and $Du(t^n,x)$. In fact $Y^n = u(t^n,X^n)$ and $Z^n = Du(t^n,X^n)$ are computed by time marching using the Euler-Maruyama scheme applied to equations \eqref{eq:2BSDE}. Their loss function is then given by \eqref{eq:loss_Arnulf} which tries to match the terminal condition. The total set of parameters are consequently given by $Y_0$, $Z_0$, $\Gamma_0$, $A_0$, and the parameters of the $2(N-1)$ neural networks used to approximate the functions $D^2u(t^n,x)$ and $\mathcal{L}Du(t^n,x)$. This framework, being a descendant of \cite{weinan2017deep,han2017overcoming}, also suffers from the drawbacks listed above. It should be emphasized that, although not pursued here, the framework proposed in the current work can be straightforwardly extended to solve the second-order backward stochastic differential equations \eqref{eq:2BSDE}. The key (see e.g., \cite{cheridito2007second}) is to leverage the fundamental relationships \eqref{eq:processes}.

\section{Results}
The proposed framework provides a universal treatment of coupled forward-backward stochastic differential equations of  fundamentally  different  nature and their corresponding high-dimensional partial differential equations. This generality  will be demonstrated by applying the algorithm to a wide range of canonical problems spanning a number of scientific domains  including a 100-dimensional Black-Scholes-Barenblatt equation and a 100-dimensional Hamilton-Jacobi-Bellman equation. These  examples are motivated by the pioneering works \cite{beck2017machine,weinan2017deep,han2017overcoming}. All data and codes used in this manuscript will be publicly available  on GitHub at \url{https://github.com/maziarraissi/FBSNNs}.

\subsection{Black-Scholes-Barenblatt Equation in 100D}

\begin{figure}[!t]
\centering
\includegraphics[scale=0.9]{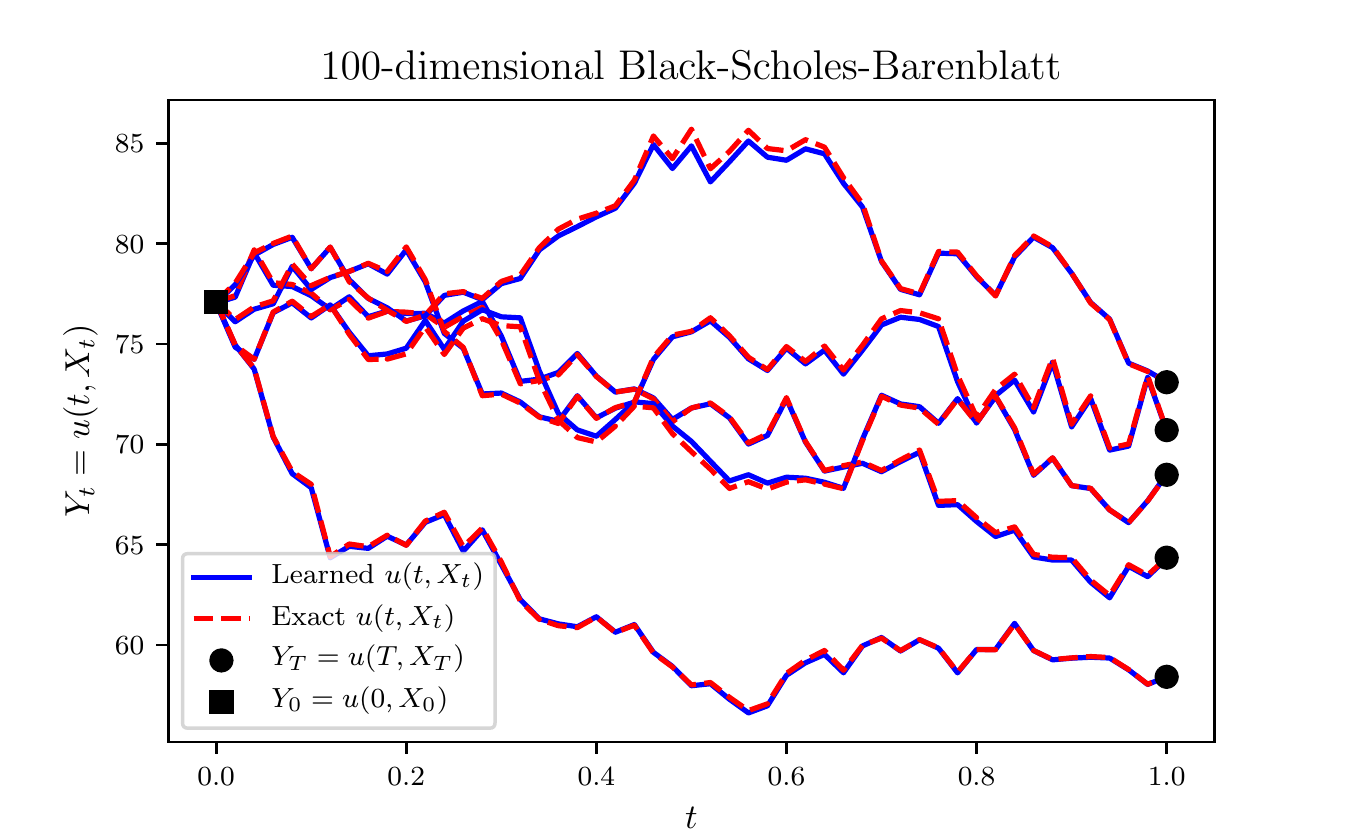}
\caption{\emph{Black-Scholes-Barenblatt Equation in 100D:} Evaluations of the learned solution $Y_t = u(t,X_t)$ at representative realizations of the underlying high-dimensional process $X_t$. It should be highlighted that the state of the art algorithms \cite{beck2017machine,weinan2017deep,han2017overcoming} can only approximate $Y_0 = u(0,X_0)$ at time $0$ and at the initial spatial point $X_0=\xi$.}\label{fig:BSB}
\end{figure}

Let us start with the following forward-backward stochastic differential equations
\begin{equation}
\begin{array}{l}
dX_t = \sigma\text{diag}(X_t)dW_t, ~~~ t \in [0,T],\\
X_0 = \xi,\\
dY_t = r(Y_t - Z_t' X_t)dt + \sigma Z_t'\text{diag}(X_t)dW_t, ~~~ t \in [0,T),\\
Y_T = g(X_T),
\end{array}
\end{equation}
where $T=1$, $\sigma = 0.4$, $r=0.05$, $\xi = (1,0.5,1,0.5,\ldots,1,0.5) \in \mathbb{R}^{100}$, and $g(x) = \Vert x \Vert ^2$. The above equations are related to the Black-Scholes-Barenblatt equation
\begin{equation}
u_t = -\frac{1}{2} \text{Tr}[\sigma^2 \text{diag}(X_t^2) D^2u] + r(u - (Du)' x),
\end{equation}
with terminal condition $u(T,x) = g(x)$. This equation admits the explicit solution
\begin{equation}
u(t,x) = \exp \left( (r + \sigma^2) (T-t) \right)g(x),
\end{equation}
which can be used to test the accuracy of the proposed algorithm. We approximate the unknown solution $u(t,x)$ by a 5-layer deep neural network with 256 neurons per hidden layer. Furthermore, we partition the time domain $[0,T]$ into $N=50$ equally spaced intervals (see equations \eqref{eq:Euler}). Upon minimizing the loss function \eqref{eq:loss}, using the Adam optimizer \cite{kingma2014adam} with mini-batches of size $100$ (i.e., 100 realizations of the underlying Brownian motion), we obtain the results reported in figure \ref{fig:BSB}. In this figure, we are evaluating the learned solution $Y_t = u(t,X_t)$ at representative realizations (not seen during training) of the underlying high-dimensional process $X_t$. Unlike the state of the art algorithms \cite{beck2017machine,weinan2017deep,han2017overcoming}, which can only approximate $Y_0 = u(0,X_0)$ at time $0$ and at the initial spatial point $X_0=\xi$, our algorithm is capable of approximating the entire solution function $u(t,x)$ in a single round of training as demonstrated in figure \ref{fig:BSB}. Figures such as this one are absent in \cite{beck2017machine,weinan2017deep,han2017overcoming}, by design.\\

\begin{figure}[!t]
\centering
\includegraphics[scale=0.9]{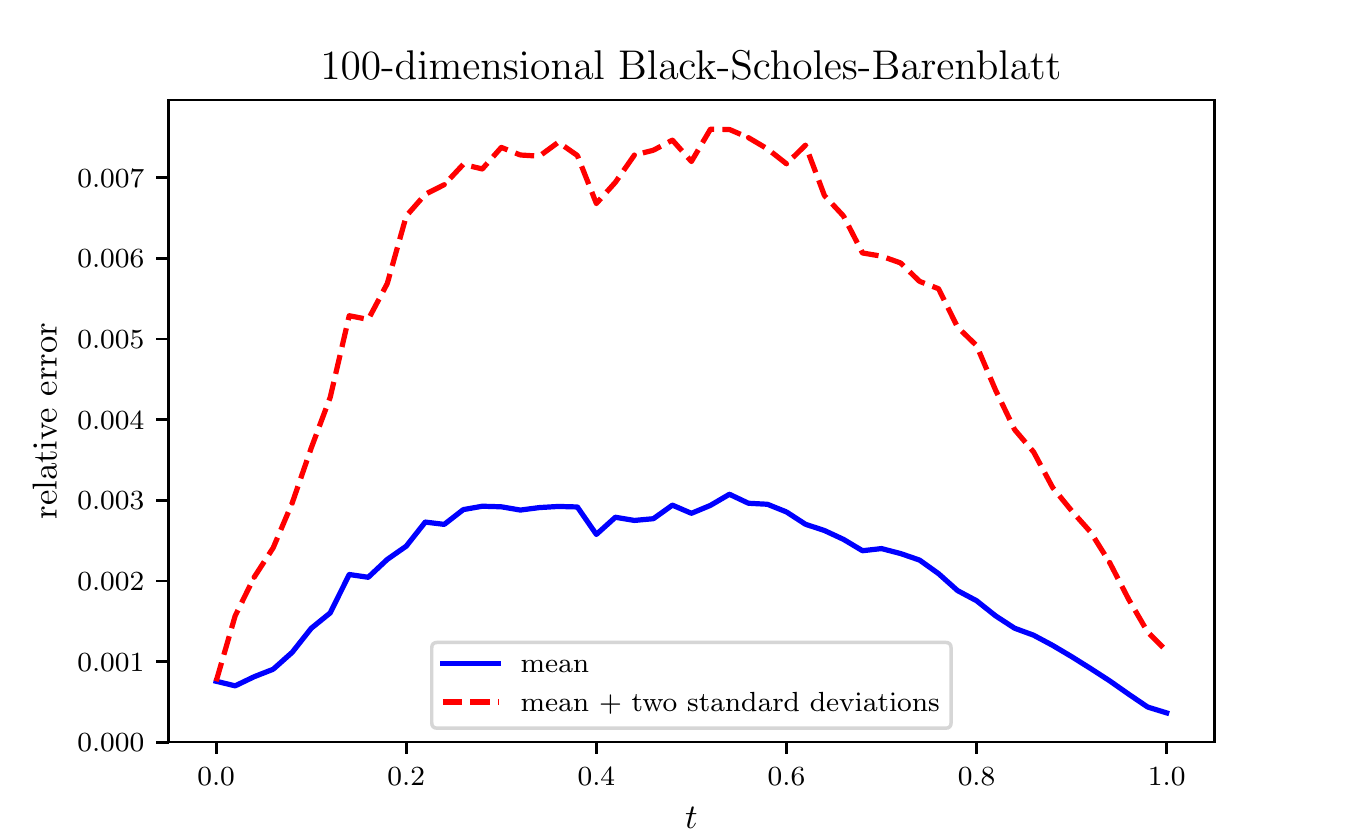}
\caption{\emph{Black-Scholes-Barenblatt Equation in 100D:} Mean and mean plus two standard deviations of the relative errors between model predictions and the exact solution computed based on $100$ realizations of the underlying Brownian motion.}\label{fig:BSB_errors}
\end{figure}

To further scrutinize the performance of our algorithm, in figure \ref{fig:BSB_errors} we report the mean and mean plus two standard deviations of the relative errors between model predictions and the exact solution computed based on $100$ independent realizations of the underlying Brownian motion. It is worth noting that in figure \ref{fig:BSB} we were plotting 5 representative examples of the 100 realizations used to generate figure \ref{fig:BSB_errors}. The results reported in figures \ref{fig:BSB} and \ref{fig:BSB_errors} are obtained after $2 \times 10^4$, $3 \times 10^4$, $3 \times 10^4$, and $2 \times 10^4$ consecutive iterations of the Adam optimizer with learning rates of $10^{-3}$, $10^{-4}$, $10^{-5}$, and $10^{-6}$, respectively. The total number of iterations is therefore given by $10^5$. Every $10$ iterations of the optimizer takes about $0.88$ seconds on a single NVIDIA Titan X GPU card. In each iteration of the Adam optimizer we are using $100$ different realizations of the underlying Brownian motion. Consequently, the total number of Brownian motion trajectories observed by the algorithm is given by $10^7$. It is worth highlighting that the algorithm converges to the exact value $Y_0 = u(0,X_0)$ in the first few hundred iterations of the Adam optimizer. For instance after only 500 steps of training, the algorithms achieves an accuracy of around $2.3 \times 10^{-3}$ in terms of relative error. This is comparable to the results reported in \cite{beck2017machine,weinan2017deep,han2017overcoming}, both in terms of accuracy and the speed of the algorithm. However, to obtain more accurate estimates for $Y_t = u(t,X_t)$ at later times $t>0$ we need to train the algorithm using more iterations of the Adam optimizer.

\subsection{Hamilton-Jacobi-Bellman Equation in 100D}

\begin{figure}[!t]
\centering
\includegraphics[scale=0.9]{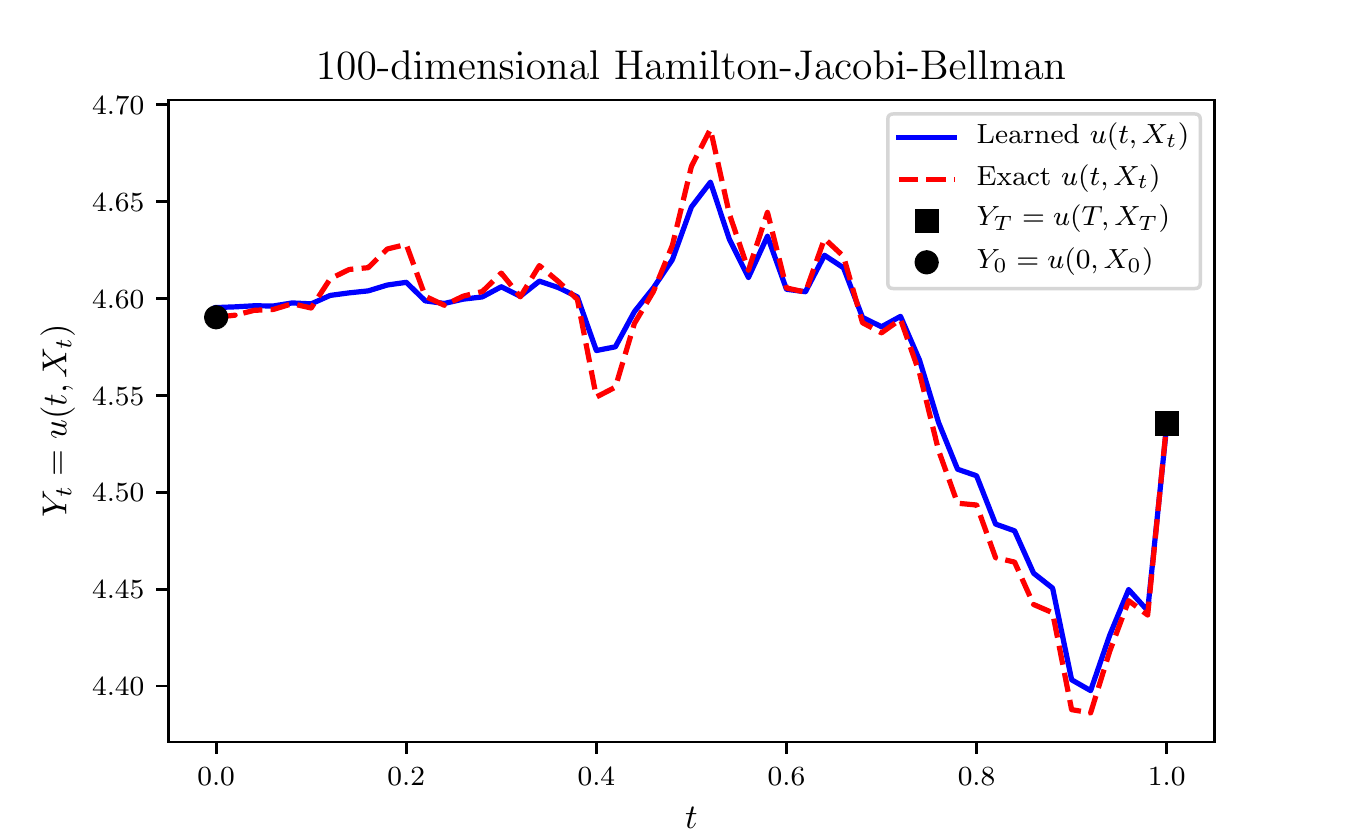}
\caption{\emph{Hamilton-Jacobi-Bellman Equation in 100D:} Evaluation of the learned solution $Y_t = u(t,X_t)$ at a representative realization of the underlying high-dimensional process $X_t$. It should be highlighted that the state of the art algorithms \cite{beck2017machine,weinan2017deep,han2017overcoming} can only approximate $Y_0 = u(0,X_0)$ at time $0$ and at the initial spatial point $X_0=\xi$.}\label{fig:HJB}
\end{figure}

Let us now consider the following forward-backward stochastic differential equations
\begin{equation}
\begin{array}{l}
dX_t = \sigma dW_t, ~~~ t \in [0,T],\\
X_0 = \xi,\\
dY_t = \Vert Z_t\Vert^2 dt + \sigma Z_t'dW_t, ~~~ t \in [0,T),\\
Y_T = g(X_T),
\end{array}
\end{equation}
where $T=1$, $\sigma = \sqrt{2}$, $\xi = (0,0,\ldots,0)\in \mathbb{R}^{100}$, and $g(x) = \ln\left(0.5\left(1+\Vert x\Vert^2\right)\right)$. The above equations are related to the Hamilton-Jacobi-Bellman equation
\begin{equation}
u_t = -\text{Tr}[D^2u] + \Vert Du\Vert^2,
\end{equation}
with terminal condition $u(T,x) = g(x)$. This equation admits the explicit solution
\begin{equation}\label{eq:HJB_exact}
u(t,x) = -\ln\left(\mathbb{E}\left[\exp\left( -g(x + \sqrt{2} W_{T-t}) \right) \right] \right),
\end{equation}
which can be used to test the accuracy of the proposed algorithm. In fact, due to the presence of the expectation operator $\mathbb{E}$ in equation \eqref{eq:HJB_exact}, we can only approximately compute the exact solution. To be precise, we use $10^5$ Monte-Carlo samples to approximate the exact solution \eqref{eq:HJB_exact} and use the result as ground truth. We represent the unknown solution $u(t,x)$ by a 5-layer deep neural network with 256 neurons per hidden layer. Furthermore, we partition the time domain $[0,T]$ into $N=50$ equally spaced intervals (see equations \eqref{eq:Euler}). Upon minimizing the loss function \eqref{eq:loss}, using the Adam optimizer \cite{kingma2014adam} with mini-batches of size $100$ (i.e., 100 realizations of the underlying Brownian motion), we obtain the results reported in figure \ref{fig:HJB}. In this figure, we are evaluating the learned solution $Y_t = u(t,X_t)$ at a representative realization (not seen during training) of the underlying high-dimensional process $X_t$. It is worth noting that computing the exact solution \eqref{eq:HJB_exact} to this problem is prohibitively costly due to the need for the aforementioned Monte-Carlo sampling strategy. That is why we are depicting only a single realization of the solution trajectories in figure \ref{fig:HJB}. Unlike the state of the art algorithms \cite{beck2017machine,weinan2017deep,han2017overcoming}, which can only approximate $Y_0 = u(0,X_0)$ at time $0$ and at the initial spatial point $X_0=\xi$, our algorithm is capable of approximating the entire solution function $u(t,x)$ in a single round of training as demonstrated in figure \ref{fig:HJB}.\\

\begin{figure}[!t]
\centering
\includegraphics[scale=0.9]{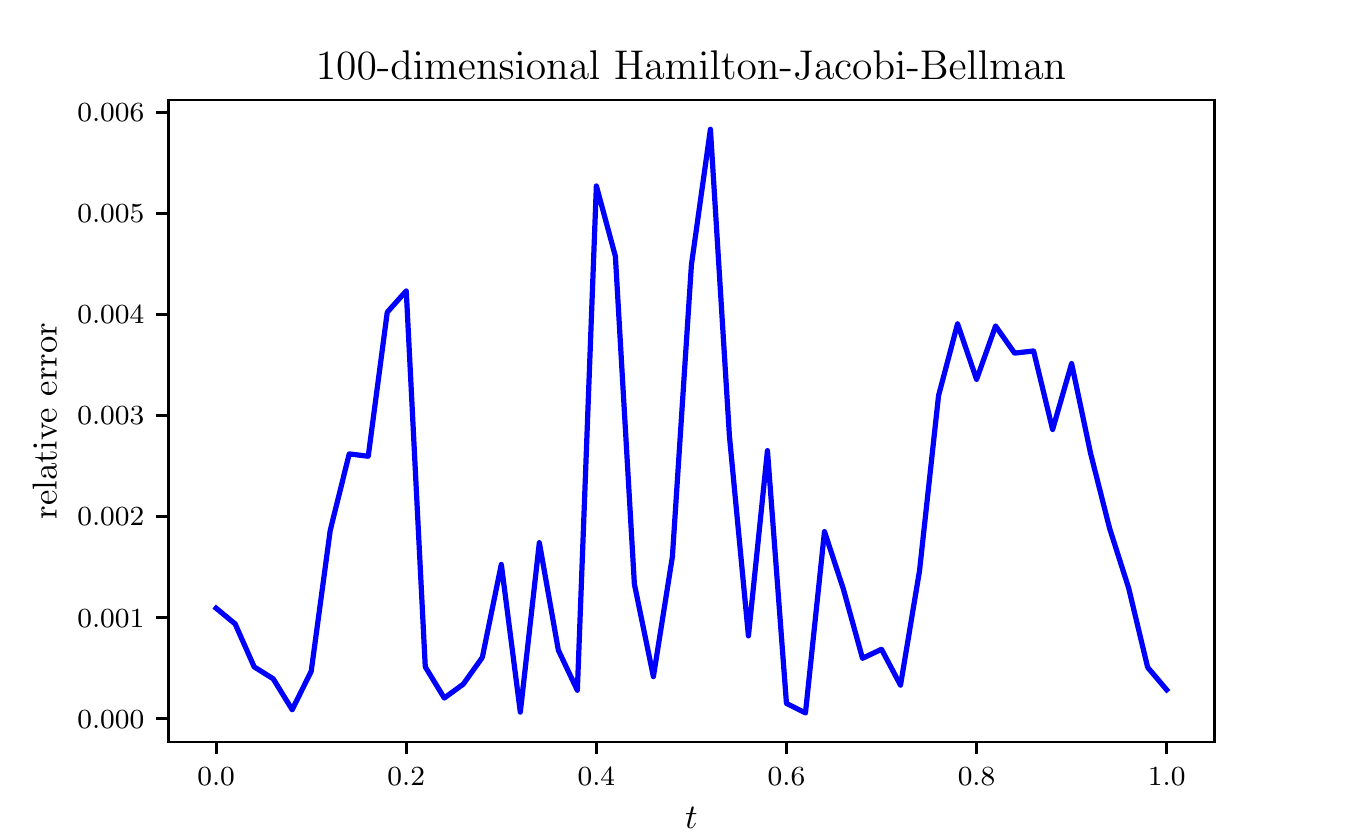}
\caption{\emph{Hamilton-Jacobi-Bellman Equation in 100D:} The relative error between model prediction and the exact solution computed based on a single realization of the underlying Brownian motion.}\label{fig:HJB_errors}
\end{figure}

To further investigate the performance of our algorithm, in figure \ref{fig:HJB_errors} we report the relative error between model prediction and the exact solution computed for the same realization of the underlying Brownian motion as the one used in figure \ref{fig:HJB}. The results reported in figures \ref{fig:HJB} and \ref{fig:HJB_errors} are obtained after $2 \times 10^4$, $3 \times 10^4$, $3 \times 10^4$, and $2 \times 10^4$ consecutive iterations of the Adam optimizer with learning rates of $10^{-3}$, $10^{-4}$, $10^{-5}$, and $10^{-6}$, respectively. The total number of iterations is therefore given by $10^5$. Every $10$ iterations of the optimizer takes about $0.79$ seconds on a single NVIDIA Titan X GPU card. In each iteration of the Adam optimizer we are using $100$ different realizations of the underlying Brownian motion. Consequently, the total number of Brownian motion trajectories observed by the algorithm is given by $10^7$. It is worth highlighting that the algorithm converges to the exact value $Y_0 = u(0,X_0)$ in the first few hundred iterations of the Adam optimizer. For instance after only 100 steps of training, the algorithms achieves an accuracy of around $7.3 \times 10^{-3}$ in terms of relative error. This is comparable to the results reported in \cite{beck2017machine,weinan2017deep,han2017overcoming}, both in terms of accuracy and the speed of the algorithm. However, to obtain more accurate estimates for $Y_t = u(t,X_t)$ at later times $t>0$ we need to train the algorithm using more iterations of the Adam optimizer.

\subsection{Allen-Cahn Equation in 20D}

\begin{figure}[!t]
\centering
\includegraphics[scale=0.9]{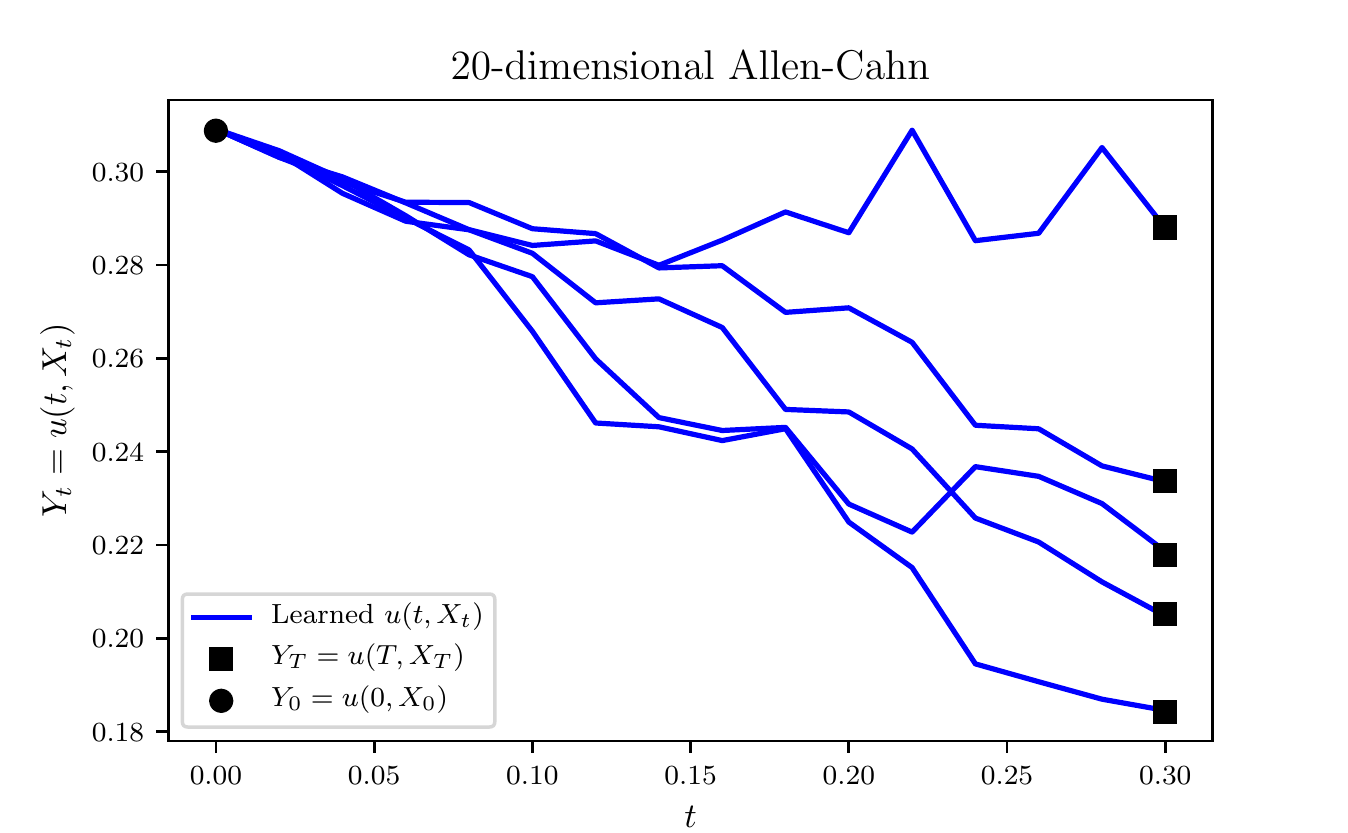}
\caption{\emph{Allen-Cahn Equation in 20D:} Evaluation of the learned solution $Y_t = u(t,X_t)$ at representative realizations of the underlying high-dimensional process $X_t$. It should be highlighted that the state of the art algorithms \cite{beck2017machine,weinan2017deep,han2017overcoming} can only approximate $Y_0 = u(0,X_0)$ at time $0$ and at the initial spatial point $X_0=\xi$.}\label{fig:AC}
\end{figure}

Let us consider the following forward-backward stochastic differential equations
\begin{equation}
\begin{array}{l}
dX_t = dW_t, ~~~ t \in [0,T],\\
X_0 = \xi,\\
dY_t = (-Y_t + Y_t^3) dt + Z_t'dW_t, ~~~ t \in [0,T),\\
Y_T = g(X_T),
\end{array}
\end{equation}
where $T=0.3$, $\xi = (0,0,\ldots,0)\in \mathbb{R}^{20}$, and $g(x) = \left(2 + 0.4\Vert x \Vert^2 \right)^{-1}$. The above equations are related to the Allen-Cahn equation
\begin{equation}
u_t = -0.5\text{Tr}[D^2u] - u + u^3,
\end{equation}
with terminal condition $u(T,x) = g(x)$. We represent the unknown solution $u(t,x)$ by a 5-layer deep neural network with 256 neurons per hidden layer. Furthermore, we partition the time domain $[0,T]$ into $N=15$ equally spaced intervals (see equations \eqref{eq:Euler}). Upon minimizing the loss function \eqref{eq:loss}, using the Adam optimizer \cite{kingma2014adam} with mini-batches of size $100$ (i.e., 100 realizations of the underlying Brownian motion), we obtain the results reported in figure \ref{fig:AC}. In this figure, we are evaluating the learned solution $Y_t = u(t,X_t)$ at five representative realizations (not seen during training) of the underlying high-dimensional process $X_t$. Unlike the state of the art algorithms \cite{beck2017machine,weinan2017deep,han2017overcoming}, which can only approximate $Y_0 = u(0,X_0) = 0.30879$ at time $0$ and at the initial spatial point $X_0=\xi$, our algorithm is capable of approximating the entire solution function $u(t,x)$ in a single round of training as demonstrated in figure \ref{fig:AC}.

\section{Summary and Discussion}
In this work, we put forth a deep learning approach for solving coupled forward-backward stochastic differential equations and their corresponding high-dimensional partial differential equations. The resulting methodology showcases a series of promising results for a diverse collection of benchmark problems. As deep learning technology is continuing to grow rapidly both in terms of methodological, algorithmic, and infrastructural developments, we believe that this is a timely contribution that can benefit practitioners across a wide range of scientific domains. Specific applications that can readily enjoy these benefits include, but are not limited to, stochastic control, theoretical economics, and mathematical finance.\\

In terms of future work, one could straightforwardly extend the proposed framework in the current work to solve second-order backward stochastic differential equations \eqref{eq:2BSDE}. The key (see e.g., \cite{cheridito2007second}) is to leverage the fundamental relationships \eqref{eq:processes} between second-order backward stochastic differential equations and fully-nonlinear second-order partial differential equations. Moreover, our method can be used to solve stochastic control problems, where in general, to obtain a candidate for an optimal control, one needs to solve a coupled forward-backward stochastic differential equation \eqref{eq:FBSDE}, where the backward components influence the dynamics of the forward component.

\section*{Acknowledgements}
This work received support by the DARPA EQUiPS grant N66001-15-2-4055 and the AFOSR grant FA9550-17-1-0013.

% \appendix
% \input{appendix.tex}


%% References


\bibliographystyle{model1-num-names}
\bibliography{sample.bib}

\begin{thebibliography}{24}
\expandafter\ifx\csname natexlab\endcsname\relax\def\natexlab#1{#1}\fi
\providecommand{\bibinfo}[2]{#2}
\ifx\xfnm\relax \def\xfnm[#1]{\unskip,\space#1}\fi
%Type = Article
\bibitem[{Bismut(1973)}]{bismut1973conjugate}
\bibinfo{author}{J.-M. Bismut},
\newblock \bibinfo{title}{Conjugate convex functions in optimal stochastic
  control},
\newblock \bibinfo{journal}{Journal of Mathematical Analysis and Applications}
  \bibinfo{volume}{44} (\bibinfo{year}{1973}) \bibinfo{pages}{384--404}.
%Type = Article
\bibitem[{Pardoux and Peng(1990)}]{pardoux1990adapted}
\bibinfo{author}{E.~Pardoux}, \bibinfo{author}{S.~Peng},
\newblock \bibinfo{title}{Adapted solution of a backward stochastic
  differential equation},
\newblock \bibinfo{journal}{Systems \& Control Letters} \bibinfo{volume}{14}
  (\bibinfo{year}{1990}) \bibinfo{pages}{55--61}.
%Type = Article
\bibitem[{Cheridito et~al.(2007)Cheridito, Soner, Touzi, and
  Victoir}]{cheridito2007second}
\bibinfo{author}{P.~Cheridito}, \bibinfo{author}{H.~M. Soner},
  \bibinfo{author}{N.~Touzi}, \bibinfo{author}{N.~Victoir},
\newblock \bibinfo{title}{Second-order backward stochastic differential
  equations and fully nonlinear parabolic pdes},
\newblock \bibinfo{journal}{Communications on Pure and Applied Mathematics}
  \bibinfo{volume}{60} (\bibinfo{year}{2007}) \bibinfo{pages}{1081--1110}.
%Type = Article
\bibitem[{Raissi et~al.(2017{\natexlab{a}})Raissi, Perdikaris, and
  Karniadakis}]{raissi2017physics_I}
\bibinfo{author}{M.~Raissi}, \bibinfo{author}{P.~Perdikaris},
  \bibinfo{author}{G.~E. Karniadakis},
\newblock \bibinfo{title}{Physics informed deep learning (part ii): Data-driven
  discovery of nonlinear partial differential equations},
\newblock \bibinfo{journal}{arXiv preprint arXiv:1711.10566}
  (\bibinfo{year}{2017}{\natexlab{a}}).
%Type = Article
\bibitem[{Raissi et~al.(2017{\natexlab{b}})Raissi, Perdikaris, and
  Karniadakis}]{raissi2017physics_II}
\bibinfo{author}{M.~Raissi}, \bibinfo{author}{P.~Perdikaris},
  \bibinfo{author}{G.~E. Karniadakis},
\newblock \bibinfo{title}{Physics informed deep learning (part i): Data-driven
  solutions of nonlinear partial differential equations},
\newblock \bibinfo{journal}{arXiv preprint arXiv:1711.10561}
  (\bibinfo{year}{2017}{\natexlab{b}}).
%Type = Article
\bibitem[{Raissi(2018)}]{raissi2018deep}
\bibinfo{author}{M.~Raissi},
\newblock \bibinfo{title}{Deep hidden physics models: Deep learning of
  nonlinear partial differential equations},
\newblock \bibinfo{journal}{arXiv preprint arXiv:1801.06637}
  (\bibinfo{year}{2018}).
%Type = Article
\bibitem[{Raissi et~al.(2018)Raissi, Perdikaris, and
  Karniadakis}]{raissi2018numerical}
\bibinfo{author}{M.~Raissi}, \bibinfo{author}{P.~Perdikaris},
  \bibinfo{author}{G.~E. Karniadakis},
\newblock \bibinfo{title}{Numerical gaussian processes for time-dependent and
  nonlinear partial differential equations},
\newblock \bibinfo{journal}{SIAM Journal on Scientific Computing}
  \bibinfo{volume}{40} (\bibinfo{year}{2018}) \bibinfo{pages}{A172--A198}.
%Type = Article
\bibitem[{Raissi and Karniadakis(2018)}]{raissi2018hidden}
\bibinfo{author}{M.~Raissi}, \bibinfo{author}{G.~E. Karniadakis},
\newblock \bibinfo{title}{Hidden physics models: Machine learning of nonlinear
  partial differential equations},
\newblock \bibinfo{journal}{Journal of Computational Physics}
  \bibinfo{volume}{357} (\bibinfo{year}{2018}) \bibinfo{pages}{125--141}.
%Type = Article
\bibitem[{Raissi et~al.(2017{\natexlab{a}})Raissi, Perdikaris, and
  Karniadakis}]{raissi2017inferring}
\bibinfo{author}{M.~Raissi}, \bibinfo{author}{P.~Perdikaris},
  \bibinfo{author}{G.~E. Karniadakis},
\newblock \bibinfo{title}{Inferring solutions of differential equations using
  noisy multi-fidelity data},
\newblock \bibinfo{journal}{Journal of Computational Physics}
  \bibinfo{volume}{335} (\bibinfo{year}{2017}{\natexlab{a}})
  \bibinfo{pages}{736--746}.
%Type = Article
\bibitem[{Raissi et~al.(2017{\natexlab{b}})Raissi, Perdikaris, and
  Karniadakis}]{raissi2017machine}
\bibinfo{author}{M.~Raissi}, \bibinfo{author}{P.~Perdikaris},
  \bibinfo{author}{G.~E. Karniadakis},
\newblock \bibinfo{title}{Machine learning of linear differential equations
  using gaussian processes},
\newblock \bibinfo{journal}{Journal of Computational Physics}
  \bibinfo{volume}{348} (\bibinfo{year}{2017}{\natexlab{b}})
  \bibinfo{pages}{683--693}.
%Type = Article
\bibitem[{Beck et~al.(2017)Beck, Jentzen et~al.}]{beck2017machine}
\bibinfo{author}{C.~Beck}, \bibinfo{author}{A.~Jentzen}, et~al.,
\newblock \bibinfo{title}{Machine learning approximation algorithms for
  high-dimensional fully nonlinear partial differential equations and
  second-order backward stochastic differential equations},
\newblock \bibinfo{journal}{arXiv preprint arXiv:1709.05963}
  (\bibinfo{year}{2017}).
%Type = Article
\bibitem[{Weinan et~al.(2017)Weinan, Han, and Jentzen}]{weinan2017deep}
\bibinfo{author}{E.~Weinan}, \bibinfo{author}{J.~Han},
  \bibinfo{author}{A.~Jentzen},
\newblock \bibinfo{title}{Deep learning-based numerical methods for
  high-dimensional parabolic partial differential equations and backward
  stochastic differential equations},
\newblock \bibinfo{journal}{Communications in Mathematics and Statistics}
  \bibinfo{volume}{5} (\bibinfo{year}{2017}) \bibinfo{pages}{349--380}.
%Type = Article
\bibitem[{Han et~al.(2017)Han, Jentzen et~al.}]{han2017overcoming}
\bibinfo{author}{J.~Han}, \bibinfo{author}{A.~Jentzen}, et~al.,
\newblock \bibinfo{title}{Overcoming the curse of dimensionality: Solving
  high-dimensional partial differential equations using deep learning},
\newblock \bibinfo{journal}{arXiv preprint arXiv:1707.02568}
  (\bibinfo{year}{2017}).
%Type = Article
\bibitem[{Antonelli(1993)}]{antonelli1993backward}
\bibinfo{author}{F.~Antonelli},
\newblock \bibinfo{title}{Backward-forward stochastic differential equations},
\newblock \bibinfo{journal}{The Annals of Applied Probability}
  (\bibinfo{year}{1993}) \bibinfo{pages}{777--793}.
%Type = Article
\bibitem[{Ma et~al.(1994)Ma, Protter, and Yong}]{ma1994solving}
\bibinfo{author}{J.~Ma}, \bibinfo{author}{P.~Protter},
  \bibinfo{author}{J.~Yong},
\newblock \bibinfo{title}{Solving forward-backward stochastic differential
  equations explicitly—a four step scheme},
\newblock \bibinfo{journal}{Probability theory and related fields}
  \bibinfo{volume}{98} (\bibinfo{year}{1994}) \bibinfo{pages}{339--359}.
%Type = Article
\bibitem[{Pardoux and Tang(1999)}]{pardoux1999forward}
\bibinfo{author}{E.~Pardoux}, \bibinfo{author}{S.~Tang},
\newblock \bibinfo{title}{Forward-backward stochastic differential equations
  and quasilinear parabolic pdes},
\newblock \bibinfo{journal}{Probability Theory and Related Fields}
  \bibinfo{volume}{114} (\bibinfo{year}{1999}) \bibinfo{pages}{123--150}.
%Type = Article
\bibitem[{Delarue and Menozzi(2006)}]{delarue2006forward}
\bibinfo{author}{F.~Delarue}, \bibinfo{author}{S.~Menozzi},
\newblock \bibinfo{title}{A forward-backward stochastic algorithm for
  quasi-linear pdes},
\newblock \bibinfo{journal}{The Annals of Applied Probability}
  (\bibinfo{year}{2006}) \bibinfo{pages}{140--184}.
%Type = Article
\bibitem[{Baydin et~al.(2015)Baydin, Pearlmutter, Radul, and
  Siskind}]{baydin2015automatic}
\bibinfo{author}{A.~G. Baydin}, \bibinfo{author}{B.~A. Pearlmutter},
  \bibinfo{author}{A.~A. Radul}, \bibinfo{author}{J.~M. Siskind},
\newblock \bibinfo{title}{Automatic differentiation in machine learning: a
  survey},
\newblock \bibinfo{journal}{arXiv preprint arXiv:1502.05767}
  (\bibinfo{year}{2015}).
%Type = Article
\bibitem[{Abadi et~al.(2016)Abadi, Agarwal, Barham, Brevdo, Chen, Citro,
  Corrado, Davis, Dean, Devin et~al.}]{abadi2016tensorflow}
\bibinfo{author}{M.~Abadi}, \bibinfo{author}{A.~Agarwal},
  \bibinfo{author}{P.~Barham}, \bibinfo{author}{E.~Brevdo},
  \bibinfo{author}{Z.~Chen}, \bibinfo{author}{C.~Citro}, \bibinfo{author}{G.~S.
  Corrado}, \bibinfo{author}{A.~Davis}, \bibinfo{author}{J.~Dean},
  \bibinfo{author}{M.~Devin}, et~al.,
\newblock \bibinfo{title}{Tensorflow: Large-scale machine learning on
  heterogeneous distributed systems},
\newblock \bibinfo{journal}{arXiv preprint arXiv:1603.04467}
  (\bibinfo{year}{2016}).
%Type = Article
\bibitem[{Raissi et~al.(2018)Raissi, Perdikaris, and
  Karniadakis}]{raissi2018multistep}
\bibinfo{author}{M.~Raissi}, \bibinfo{author}{P.~Perdikaris},
  \bibinfo{author}{G.~E. Karniadakis},
\newblock \bibinfo{title}{Multistep neural networks for data-driven discovery
  of nonlinear dynamical systems},
\newblock \bibinfo{journal}{arXiv preprint arXiv:1801.01236}
  (\bibinfo{year}{2018}).
%Type = Article
\bibitem[{Raissi(2017)}]{raissi2017parametric}
\bibinfo{author}{M.~Raissi},
\newblock \bibinfo{title}{Parametric gaussian process regression for big data},
\newblock \bibinfo{journal}{arXiv preprint arXiv:1704.03144}
  (\bibinfo{year}{2017}).
%Type = Article
\bibitem[{Perdikaris et~al.(2017)Perdikaris, Raissi, Damianou, Lawrence, and
  Karniadakis}]{perdikaris2017nonlinear}
\bibinfo{author}{P.~Perdikaris}, \bibinfo{author}{M.~Raissi},
  \bibinfo{author}{A.~Damianou}, \bibinfo{author}{N.~Lawrence},
  \bibinfo{author}{G.~E. Karniadakis},
\newblock \bibinfo{title}{Nonlinear information fusion algorithms for
  data-efficient multi-fidelity modelling},
\newblock \bibinfo{journal}{Proc. R. Soc. A} \bibinfo{volume}{473}
  (\bibinfo{year}{2017}) \bibinfo{pages}{20160751}.
%Type = Article
\bibitem[{Raissi and Karniadakis(2016)}]{raissi2016deep}
\bibinfo{author}{M.~Raissi}, \bibinfo{author}{G.~Karniadakis},
\newblock \bibinfo{title}{Deep multi-fidelity {G}aussian processes},
\newblock \bibinfo{journal}{arXiv preprint arXiv:1604.07484}
  (\bibinfo{year}{2016}).
%Type = Article
\bibitem[{Kingma and Ba(2014)}]{kingma2014adam}
\bibinfo{author}{D.~P. Kingma}, \bibinfo{author}{J.~Ba},
\newblock \bibinfo{title}{Adam: A method for stochastic optimization},
\newblock \bibinfo{journal}{arXiv preprint arXiv:1412.6980}
  (\bibinfo{year}{2014}).

\end{thebibliography}






% \end{thebibliography}

\end{document}